\pdfoutput=1

\documentclass[11pt]{article}

\usepackage[]{ACL2023}

\usepackage{times}
\usepackage{latexsym}

\usepackage[T1]{fontenc}

\usepackage[utf8]{inputenc}

\usepackage{microtype}

\usepackage{inconsolata}
\usepackage{graphicx} 
\usepackage{booktabs}
\usepackage{amsmath}
\usepackage{multirow}
\title{Enhancing Open-Domain Table Question Answering via Syntax- and Structure-aware Dense Retrieval}

\author{Nengzheng Jin$^{1}$, Dongfang Li$^{1}$, Junying Chen$^{1}$, Joanna Siebert$^{1}$, Qingcai Chen$^{1,2}$\\ $^{1}$Harbin Institute of Technology (Shenzhen), Shenzhen, China \\ $^{2}$Peng Cheng Laboratory\\
  \texttt{\{nengzhengjin,crazyofapple,junying.chen.cs,joannasiebert\}@gmail.com}\\\texttt{qingcai.chen@hit.edu.cn}}

\begin{document}
\maketitle


\begin{abstract}

Open-domain table question answering aims to provide answers to a question by retrieving and extracting information from a large collection of tables. Existing studies of open-domain table QA either directly adopt text retrieval methods or consider the table structure only in the encoding layer for table retrieval, which may cause syntactical and structural information loss during table scoring.  To address this issue, we propose a syntax- and structure-aware retrieval method for the open-domain table QA task. It provides syntactical representations for the question and uses the structural header and value representations for the tables to avoid the loss of fine-grained syntactical and structural information. Then, a syntactical-to-structural aggregator is used to obtain the matching score between the question and a candidate table by mimicking the human retrieval process. Experimental results show that our method achieves the state-of-the-art on the NQ-tables dataset and overwhelms strong baselines on a newly curated open-domain Text-to-SQL dataset\footnote{The data and processing code are publicly available at https://github.com/nzjin/ODTQA.}. 
\end{abstract}

\section{Introduction}

Open-domain table QA uncovers the necessity of table retrieval for practical applications. It slightly differs from most table QA tasks (e.g., table semantic parsing \cite{DBLP:conf/emnlp/YinN18, DBLP:conf/acl/WangSLPR20} and end-to-end table QA \cite{DBLP:conf/emnlp/MullerPSNA19,DBLP:conf/emnlp/EisenschlosG0C21}), which typically assume that the relevant tables of a question are provided at test time. This assumption can hardly hold when the user is asking questions through some open-domain natural language interface or querying large databases. 

Hence, some recent studies \cite{herzig-etal-2021-open,Chen2022RetrievalAV,DBLP:journals/corr/abs-2201-05880} have started to explore open-domain table QA. Generally, these methods are based on a two-stage framework, namely the retriever-reader (parser) framework. In such frameworks, the retriever is utilized to obtain relevant tables of the given question and the reader (parser) provides answers to the question directly or parses the question into an SQL query. 
 
Since retrieval is the first key factor for open table QA, recent works have investigated different approaches for table retrieval, which could be broadly classified into two categories. One is directly adopting text retrieval methods for table retrieval. Such methods typically linearize the tables into text and apply sparse text retrieval (e.g., BM25) \cite{DBLP:conf/acl/LiNXZWX20} or dense passage retrieval with a Bi-encoder model \cite{DBLP:conf/naacl/OguzCKPOSGMY22,DBLP:conf/iclr/ChenCSWC21,DBLP:conf/emnlp/HuangZLGJD22}. Another direction is to follow the framework of text retrieval models while also considering the unique characteristics of tables. For instance, DTR \cite{herzig-etal-2021-open} follows the Bi-encoder framework but incorporates row/column features into the table encoder, aiming to specify the cell location and enhance table understanding. UTP \cite{UTP} introduces pre-training and cross-model contrastive regularization for better tabular understanding. 

\begin{figure}
    \centering
    \includegraphics[width=0.47\textwidth]{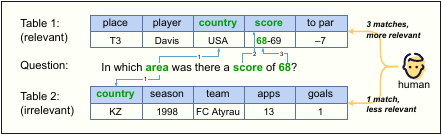}
    \caption{Illustration of fine-grained semantic matching when a human retrieves relevant tables for a question. If more fine-grained matches are found in a table, one will treat the table as more relevant.}
    \label{fig:1}
\end{figure}

Although some of these methods \cite{herzig-etal-2021-open,UTP} have modeled the characteristics of tables, they still have two shortages. First, the learned tabular semantics may be compromised when all token embeddings are combined into a single table representation in the Bi-encoder retrieval framework \cite{DBLP:journals/corr/abs-1811-08008Biencoder}. Second, they neglect that table retrieval is a fine-grained semantic matching problem. As illustrated in Figure \ref{fig:1}, humans would consciously match each meaningful phrase in the question to the table columns and rank the candidate tables by matching degree.

Motivated by this, we propose a syntax- and structural-aware dense retrieval method to mimic this fine-grained matching process. We first apply syntactic analysis to extract all possible meaningful phrases. Then, a corresponding syntactical representation is generated for each meaningful phrase based on the phrase token embeddings. To obtain fine-grained structural representations, we provide one representation for each table header. Further, we observe that the semantics of a table header and its value may be different; for example, a header may be "age" but the corresponding values are numbers. Therefore, we also provide one representation for the values of a column to retain the structural semantics better. Finally, the matching score between the question and a candidate table is obtained by performing syntactical-to-structural aggregation over the fine-grained representations, wherein the aggregation is analogous to the human behavior of counting matches.

\section{Methodology}

\subsection{Overview}

\begin{figure}
    \centering
    \includegraphics[width=0.45\textwidth]{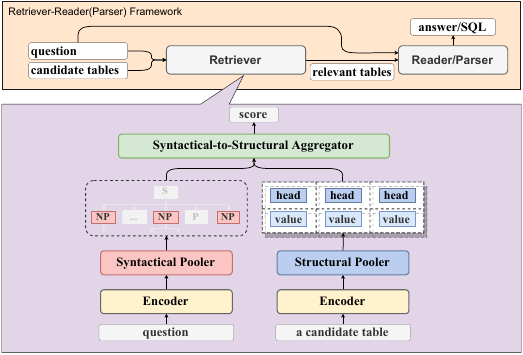}
    \caption{Illustration of the retriever-reader (parser) framework and our proposed retriever. Here, \textit{head} and \textit{value} denote the structural table representations. \textit{NP} (noun phrase) denotes a syntactical question representation. }
    \label{fig:model}
\end{figure}

As shown in Figure \ref{fig:model}, our proposed retrieval model comprises of 1) \textit{Syntactical representation module} that generates fine-grained syntactical representations for the question; 2) \textit{Structural representation module} that generates a limited number of structural representations for a table; 3) \textit{Syntactical-to-structural aggregator} that produces the matching score between the question and a candidate table.

\subsection{Syntactical Question Representation}

\ \ \ \ \textit{\textbf{Explicit syntactical representations}}. We utilize an explicit syntax parser from the Natural Language Toolkit (NLTK) \cite{NLTK} to extract the noun phrases in the question. To generate the question representations, we first feed the question tokens of length $L$ into the encoder and treat the hidden states of the last layer as question token embeddings $\{\mathbf{h}_l\}_{l=1}^L$. Then, we apply a mean pooling function over the token embeddings that belong to a phrase, thus obtaining one representation $\mathbf{q}_i$ for the $i$th phrase in the question:
\begin{equation}
    \mathbf{q}_i = \operatorname{Pooling}(\mathbf{h}_{start_i},...,\mathbf{h}_{end_i}),
\end{equation}
where $start_i$ and $end_i$ denote the token span of the $i$th phrase. Applying the same operation for each phrase, we obtain a small group of fine-grained syntactical representations $\{\mathbf{q}_i\}_{i=1}^n$, where $n$ is the number of the syntactical representations.

\textit{\textbf{Implicit syntactical representations}}. Since explicit syntax parsing is associated with additional preprocessing time, we provide an implicit syntactical representation approach that does not require syntax parsing. It attempts to learn a static number of syntactical representations through an attentive learning mechanism based on the token embeddings $\{\mathbf{h}_l\}_{l=1}^L$. Specifically, we first define a set of randomly initialized learning embeddings $\{\mathbf{a}_i\}_{i=1}^n$. Then, we feed $\{\mathbf{a}_i\}_{i=1}^n$ and $\{\mathbf{h}_l\}_{l=1}^L$ into an attention layer to generate the final syntactical representations $\{\mathbf{q}_i\}_{i=1}^n$, where $\mathbf{q}_i$ is learned as in Eqn. \ref{question_repr_1} and Eqn. \ref{question_repr_2}.
\begin{equation}
    \mathbf{q}_i = \sum_l \mathbf{w}_l^{i} \mathbf{h}_l 
    \label{question_repr_1} 
\end{equation}
\begin{equation}
    (\mathbf{w}_1^{i},.., \mathbf{w}_L^{i}) = \operatorname{softmax}(\mathbf{a}_i \cdot \mathbf{h}_1,..,\mathbf{a}_i \cdot \mathbf{h}_L).
    \label{question_repr_2}
\end{equation}

Here, $\mathbf{a}_i$ is used as a seed to learn an implicit syntactical representation $\mathbf{q}_i$ during training.
\subsection{Structural Table Representation}

For each candidate table, we linearize the table into a sequence by concatenating the columns, wherein a column sequence consists of the column name (also referred to as the header) and column values. Then, we use the same encoder to encode the table sequence of length $T$. After obtaining table token embeddings $\{\mathbf{s}_l\}_{l=1}^T$, we apply a mean pooling function over the embeddings of each column name to generate a header representation $\textbf{c}_{j}^{head}$. Additionally, we perform the same pooling function over the column value to generate a value representation $\textbf{c}_{j}^{val}$. The pooling is only applied to the first column value rather than all values as it proved to be more effective and efficient in our preliminary experiments. After that, we obtain two representations for the $j$th column as follows:
\begin{equation}
    \textbf{c}_{j}^{head} = \operatorname{Pooling}(\textbf{s}_{start_j^{head}},..,\textbf{s}_{end_j^{head}}),
    \label{table_repr_1}
\end{equation}
\begin{equation}
    \textbf{c}_{j}^{val} = \operatorname{Pooling}(\textbf{s}_{start_j^{val}},..,\textbf{s}_{end_j^{val}}).
    \label{table_repr_2}
\end{equation}

Here, $start_j^{head/val}$ and $end_j^{head/val}$ represent the token span corresponding to the header (value) of the $j$th column.

\subsection{Syntactical-to-Structural Aggregator}

We obtain syntactical question representations $\{\mathbf{q}_i\}_{i=1}^n$ and structural table representations $\{\mathbf{c}_j\}_{j=1}^{2m}$ after representation generation, where $\{\mathbf{c}_j\}_{j=1}^{2m}$ includes $\{\mathbf{c}_j^{head}\}_{j=1}^m$ and $\{\mathbf{c}_j^{val}\}_{j=1}^m$, and $m$ is the number of columns. Then, we perform a \textit{maxsim} operation \cite{DBLP:conf/sigir/KhattabZ20,DBLP:journals/tacl/LuanETC21} over the syntactical and structural representations to retrieve tables. Specifically, we calculate the dot product similarity between each syntactical representation $\mathbf{q}_i$ and each structural representation $\mathbf{c}_j$ to obtain a fine-grained matching score $w_{ij}$ as follows: 

\begin{equation}
    w_{ij} = \mathbf{q}_i \cdot \mathbf{c}_j
    \label{attention_agg_1}
\end{equation}
\begin{equation}
    Score = \sum_i^{n} \max_{j \in [1,2m]} w_{ij}.
    \label{attention_agg_2}
\end{equation}

Subsequently, we select the fine-grained score of the most matched column for each syntactical representation and sum up all the greatest fine-grained scores as the final matching score between the question and a candidate table. This is analogous to the process of a human finding a match for each phrase and counting the number of matches when retrieving tables.

\section{Experiments}

\subsection{Experimental Data and Settings}

\ \ \ \ \textbf{\textit{Datasets}}. We conducted experiments on two datasets: \textbf{NQ-TABLES} \cite{herzig-etal-2021-open}, \textbf{WikiSQL} \cite{DBLP:journals/corr/Seq2SQL}. NQ-TABLES is an open-domain table QA database constructed from the Natural Question dataset \cite{NQ}. In the original WikiSQL dataset, the relevant tables are given by humans. To simulate a realistic situation, we remove the table labeling of the questions and introduce table retrieval. Further details of the dataset modification can be seen in Appendix \ref{dataset construction appendix}. Concurrently with this work, \cite{open-wikitables} also introduce an open-domain setting to the WikiSQL dataset but does not process those same-header database tables.

\textbf{\textit{Settings}}.  The uncased BERT-base \cite{DBLP:conf/naacl/DevlinCLT19BERT} is employed as the encoder. The number of implicit syntactical representations is set to 3 and Adam is utilized \cite{DBLP:journals/corr/KingmaB14} as the optimizer. To evaluate the performance of the retrievers, the recall@K and exact match accuracy (EM) of the final answers are utilized as the metrics.  Since the focus of this work is to enhance open-domain table QA via improved retrievers, TAPAS \cite{DBLP:conf/acl/HerzigNMPE20} and HydraNet \cite{Lyu2020HybridRN} are directly used as the reader (parser) on the NQ-TABLES and WikiSQL datasets, respectively.

\begin{table}[t]
    \centering
    \small
    \begin{tabular}{l@{\hspace{3.5\tabcolsep}}c@{\hspace{2.5\tabcolsep}}c@{\hspace{2.5\tabcolsep}}c@{\hspace{2.5\tabcolsep}}c}
        \toprule
        Model & R@1 & R@10 & R@50 & EM \\
        \midrule
        \text{BM25}& 16.77 & 40.06 & 58.39 & 21.46  \\
        \text{DTR-Schema}& 34.36 & 74.24 & 88.37 & 32.75 \\
        \text{DTR}& 36.24 & 76.02 & 90.25 & 35.50 \\
        UTP & 38.45 & 79.03 & 92.21 & - \\
        Ours(ex) & 45.15 & 83.73 & 93.12 & 37.73 \\
        Ours(im) & \textbf{47.03} & \textbf{84.76} & \textbf{94.89} &  \textbf{37.98} \\
        
        \midrule
        \multicolumn{5}{l}{With hard negatives training} \\
        DTR & 42.42 & 81.13 & 92.56 & 37.69 \\
        UTP & 50.39 & 85.40 & 94.31 & - \\
        Tri-encoder & - & 86.4 & - & - \\
        Ours(ex) & 53.39 & 88.11 & 95.09 & 39.47 \\
        Ours(im) & \textbf{54.12} & \textbf{90.41} & \textbf{97.18} &  \textbf{39.72}\\
        \bottomrule
    \end{tabular}
    \caption{Experimental results on the NQ-TABLES dataset. R is short for recall. Here, ex/im denotes using explicit or implicit syntactical representations.}
    \label{tab:main_experiment_1}
\end{table}

\subsection{Comparison Models}
We compared the proposed retriever with the following baselines. 1) Sparse retrievers: \textbf{TF-IDF} and \textbf{BM25} \cite{DBLP:conf/sigir/BM25}.
2) Dense retrievers using one representation, such as \textbf{Bi-encoder} \cite{DBLP:journals/corr/abs-1811-08008Biencoder} with representation vectors from word2vec (w2v) and BERT, as well as \textbf{DTR} \cite{herzig-etal-2021-open} and \textbf{UTP} \cite{UTP}, which use a table-oriented pre-trained model as encoder. 3) Dense retrievers using multiple representations: \textbf{Tri-encoder} \cite{tri-encoder}, \textbf{PolyEncoder} \cite{DBLP:conf/iclr/HumeauSLW20} and \textbf{MEBERT} \cite{DBLP:journals/tacl/LuanETC21}. We set the representation number of PolyEncoder and MEBERT to the average representation number used in our model.

\subsection{Experimental Results and Analysis}

Table \ref{tab:main_experiment_1} demonstrates the experimental results on the NQ-TABLES dataset. It shows that our proposed model outperforms the previous state-of-the-art model by a considerable margin with or without hard negative training \cite{DBLP:conf/conll/GillickKLPBIG19}. Specifically, our model surpasses previous models by approximately 8 and 2 points in terms of recall@1 and EM accuracy. Furthermore, as Table \ref{tab:main_experiment_2} illustrates, our method consistently outperforms strong text retrievers on the WikiSQL dataset. This verifies the effectiveness of the proposed syntax- and structure-aware dense retrieval method.

\begin{table}[t]
    \centering
    \small
    \begin{tabular}{lcccc}
        \toprule
        Model & R@1 & R@5 & R@20 & EM \\
        \midrule
        TF-IDF & 8.93 & 22.86 & 44.05 & 7.24\\
        $\text{BM25}$& 28.19 & 44.71 & 61.22 & 26.18 \\
        Bi-encoder(w2v) & 10.01 & 20.14 & 30.13 & 7.92\\
        Bi-encoder(BERT) & 48.89 & 73.15 & 86.64 & 41.26 \\
        MEBERT & 49.05 & 73.30 & 87.34 & 41.32 \\
        PolyEncoder & 49.33 & 73.61 & 87.78 & 41.63 \\
        \midrule
        Ours(ex) & \textbf{54.16} & \textbf{77.63} & 89.81 & \textbf{45.42} \\
        Ours(im) & 53.19 & 77.57 & \textbf{90.02} & 44.71 \\
        \bottomrule
    \end{tabular}
    \caption{Experimental results on the WikiSQL dataset.}
    \label{tab:main_experiment_2}
\end{table}

The experimental results in Table \ref{tab:main_experiment_1} and \ref{tab:main_experiment_2} also indicate that the implicit syntactical question representations yield better performance compared to the explicit ones in most cases. The underlying reason may be that the attention layer in the implicit syntactical pooler can learn more delicate semantic information compared to simple pooling. 

\textbf{Ablation Study}. To analyze the impact of different components of our proposed method, we conducted an ablation study in Table \ref{tab:ablation}. 1) w/o  $\mathcal{S}^1$ variant eliminates syntactical question representations and takes the embedding of the $[CLS]$ token in the question as the representation. The result of w/o  $\mathcal{S}^1$ shows the importance of syntactical representations. Moreover, our method still outperforms text retrieval baselines in this case, which also verifies the effectiveness of structural table representations compared with multiple contextualized representations \cite{DBLP:conf/iclr/HumeauSLW20, luan-etal-2021-sparse}.  2) w/o $\mathcal{S}^2$ variant uses the embedding of the $[CLS]$ token in the table sequence as the table representation. The results of w/o  $\mathcal{S}^2$, w/o  $\mathcal{S}^2 (head)$, and w/o  $\mathcal{S}^2 (value)$ indicate that structural header and value representations both contribute to better representations of a table. 3) w/o  $\mathcal{S}^1 + \mathcal{S}^2$ variant leads to inferior performance but is slightly better than the w/o  $\mathcal{S}^2$ variant, this suggests that the use of syntactical representations alone does not improve performance.

\begin{table}[t]
    \centering
    \small
    \begin{tabular}{lcccc}
        \toprule
        Model & R@1 & R@5 & R@20 & EM \\ 
        \midrule
        Ours(ex) &  54.16 & 77.63 & 89.81 & 45.42 \\
        w/o $\mathcal{S}^1$ & 50.08& 74.36 & 88.19 & 42.28\\
        w/o $\mathcal{S}^2\ (head)$ & 52.93 & 77.08 & 89.91 & 44.51\\
        w/o $\mathcal{S}^2\ (value)$ & 53.18 & 77.16 & 89.50 & 44.75\\
        w/o $\mathcal{S}^2$ & 48.31 & 72.45 & 86.75 &  40.71\\
        w/o $\mathcal{S}^1+\mathcal{S}^2$ & 48.87 & 72.57 & 86.79 & 41.17\\
        \bottomrule
    \end{tabular}
    \caption{Ablation sturdy on the WikiSQL dataset. $\mathcal{S}^1$ and $\mathcal{S}^2$ denote syntactical and structural representations.}
    \label{tab:ablation}
\end{table}

\begin{table}[t]
    \centering
    \small
    \label{tab:my_label}
    \begin{tabular}{l@{\hspace{0.6\tabcolsep}}|@{\hspace{0.6\tabcolsep}}c@{\hspace{0.6\tabcolsep}}|@{\hspace{0.6\tabcolsep}}l@{\hspace{0.6\tabcolsep}}|@{\hspace{0.6\tabcolsep}}c}
    \toprule
        Model & LAT. & Model & LAT. \\
    \midrule
         TF-IDF& 2.74 & MEBERT & 0.40 \\
         BM25& 6.95 & PolyEncoder & 0.41\\
         Bi-encoder(w2v)& 0.32 & Ours(ex)& $0.42+0.6^{\ast}$ \\
         Bi-encoder(BERT) & 0.36 & Ours(im) & 0.47 \\
    \bottomrule
    \end{tabular}
    \caption{The retrieval latency (LAT.) per question (in milliseconds) on the WikiSQL dataset. ${\ast}$ denotes the latency of syntax parsing.}
    \label{tab:latency}
\end{table}

\textbf{Latency Analysis}. We compared the retrieval latency of the retrievers on the WikiSQL dataset. As Table \ref{tab:latency} illustrates, our model has an acceptable latency increase compared to the baselines but makes considerable progress in retrieval performance.

\section{Related Works}

Open-domain table QA (ODTQA) is an extension of the closed-domain table QA task \cite{DBLP:conf/acl/PasupatL15WTQ, DBLP:journals/corr/Seq2SQL, DBLP:conf/acl/YinNYR20, DBLP:conf/iclr/ChenCSWC21}. Traditional closed-domain table QA does not require table retrieval and can be addressed through two primary methodologies: employing end-to-end models or utilizing semantic parsing approaches. End-to-end models take both the questions and relevant tables as input, and then directly generate answers \cite{DBLP:conf/emnlp/MullerPSNA19, DBLP:conf/acl/ZhuLHWZLFC20TATQA, DBLP:conf/ijcnlp/NararatwongKI22, nan-etal-2022-fetaqa}. Differently, semantic parsing approaches transform the question into a logical form (e.g., a SQL query), and then retrieve the answers by executing the logical form \cite{DBLP:journals/corr/Seq2SQL, DBLP:conf/emnlp/Yuspider18, DBLP:conf/acl/WangSLPR20,DBLP:conf/acl/ZhuLHWZLFC20TATQA}.

ODTQA is also closely related to open-domain text-based QA \cite{NQ, Khashabi2021GooAQOQ} and information retrieval \cite{DBLP:journals/tacl/LuanETC21, DBLP:conf/iclr/HumeauSLW20,DBLP:conf/acl/TangSJWZW20}. Compared to the text retrieval, the tabular characteristics need to be considered in ODTQA \cite{herzig-etal-2021-open, UTP, tri-encoder}. Another task that ODTQA shares some degree of similarity with is keyword-based web table search \cite{DBLP:conf/www/zhangshuoSTR, DBLP:journals/ijon/SunYTDQ19,DBLP:conf/sigir/ChenTHX020,DBLP:conf/sigir/00600CPS21,DBLP:conf/www/TrabelsiC00H22}. Compared with keyword-based web table search, table retrieval in open-domain table QA needs to process complex questions that may contain superfluous information rather than to process informative keyword queries. Hence, our work incorporates syntax analysis to extract useful syntactical representation, as well as uses simple yet effective structural representation and scoring mechanism for retrieval efficiency.

\section{Conclusion}

In this paper, we present a syntax- and structure-aware dense retrieval method for open-domain table QA. Specifically, our method mimics the human retrieval process by utilizing fine-grained syntactical and structural representations, as well as a syntactical-to-structural aggregator. Experimental results on two datasets demonstrate that our model surpasses the strong baselines while preserving a reasonable computation overhead.

\bibliography{anthology,custom}
\bibliographystyle{acl_natbib}

\newpage

\appendix

\section{Dataset Modification}
\label{dataset construction appendix}

In the original WikiSQL dataset, tables relevant to each question are manually provided. To simulate a realistic situation, we removed the table labeling of the questions, introducing the task of table retrieval. Moreover, we merged the tables with the same headers into one distinct table, as it is implausible for an application database to accommodate two tables with identical headers. Furthermore, we created a mapping between the original tables and the distinct tables to enable using all samples for table retrieval and downstream tasks. 

In consideration of the limited input length of a bert-base encoder, we randomly sampled 5 rows of values for over-size tables, thereby obtaining smaller candidate tables during retrieval.  If the resulting table sequence still exceeded the prescribed length limitation, we dynamically trimmed an equal number of values from each column until the specified limitation was met. The statistics of the processed WikiSQL dataset and the original NQ-TABLES dataset are reported in Table \ref{tab:dataset_statistic}.

\begin{table}[!htb]
    \centering
    \small
    \begin{tabular}{lccccc}
        \toprule
        Dataset & Train & Dev & Test  & \# Tables  \\
        \midrule
        NQ-TABLES & 9594 & 1068 & 966 & 169898  \\
        WikiSQL & 56355 & 8421 & 15878 & 9898 \\
        \bottomrule
    \end{tabular}
    \caption{Statistics of NQ-TABLES and WikiSQL datasets.}
    \label{tab:dataset_statistic}
\end{table}

\section{Experimental Setup}

\label{experimental setup appendix}

Our proposed retrieval model uses a training batch size of 144 and 128 on the NQ-TABLES dataset and WikiSQL dataset, respectively. The uncased BERT-base model is used as the encoder on two datasets. Following Tri-encoder \cite{tri-encoder} and UTP \cite{UTP}, no down-projection is used for the final representations on the NQ-TABLES dataset. However, for a comprehensive evaluation, we also reported the performance of the proposed model with projection in Table \ref{dim}. We trained the model for a maximum of 150 epochs on two datasets and adopted early stopping according to the recall numbers on the dev set. For efficiency, we only used the tables that appear in the dev set as the candidate pool for the early stopping, aligning with the method outlined in \cite{herzig-etal-2021-open}.  

We trained all models on 4 Nvidia Tesla A100 80GB GPUs. We tested each model on a single GPU. Detailed hyper-parameters for training the proposed retrieval model can be found in Table \ref{hyper-parameter}.

\begin{table}[!htb]
    \centering
    \small
    \begin{tabular}{@{\hspace{2\tabcolsep}}l@{\hspace{3.5\tabcolsep}}|@{\hspace{3.5\tabcolsep}}c@{\hspace{2\tabcolsep}}}
    \toprule
         parameter& value  \\
         \midrule
         learning rate & 2e-5 \\
         weight decay & 0.01 \\
         warmup up ratio & 0.05 \\
         \bottomrule
    \end{tabular}
    \caption{Hyper-parameters for training the proposed retrieval model.}
    \label{hyper-parameter}
\end{table}

\begin{table}[!htb]
    \centering
    \small
    \begin{tabular}{l@{\hspace{0.9\tabcolsep}}c@{\hspace{0.9\tabcolsep}}c@{\hspace{0.9\tabcolsep}}cccc}
        \toprule
         Model & Size & Dim. & Bs. & R@1 & R@10 & R@50 \\
         \midrule
         DTR & large & 256 & 256 & 36.24 & 76.02 & 90.25  \\
         Ours(ex) & base & 256 & 144 & 41.29 & 82.06 & 91.86 \\
         Ours(im) & base & 256 & 144 &  41.92 & 83.42 & 93.84 \\
         \bottomrule
    \end{tabular}
    \caption{Experimental results of the proposed method with down-projection on the NQ-TABLES dataset. Here, Size denotes the model size, Dim. denotes the dimensionality of each representation after down-projection, Bs. denotes the training batch size.}
    \label{dim}
\end{table}

\section{Further Discussion}
\begin{figure*}[t]
    \centering
    \includegraphics[width=0.83\textwidth]{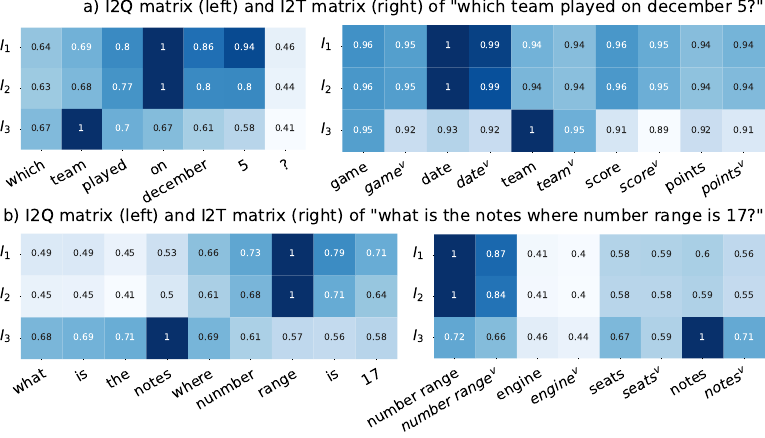}
    \caption{The coherence matrices of \textbf{I}mplicit syntactical representations ($\{I_j\}_{j=1}^3$) to \textbf{Q}uestion token embeddings (\textbf{I2Q}) and \textbf{T}able header/value representations (\textbf{I2T}). The matrices depict whether $I_j$ learns syntactical information and matches the correct column. As shown in a), $I_1$ and $I_2$ learn information of "on December 5" and match the column "date" with the highest score. Here, $\mathbf{header^v}$ denotes value representations.}
    \label{fig:attention_visualization}
\end{figure*}

\subsection{Impact of Different Pooling Functions on Performance}

In the main experiments, we used the \textbf{mean pooling function} to retrieve explicit syntactical and structural representations. This function outputs the average of all input embeddings. Here, we explored the effect of two other pooling functions, namely the \textbf{max pooling function}, which selects the largest of the input embeddings as the output, and the \textbf{attentive pooling function}, which generates the output as a weighted sum of the input embeddings, with weights determined by a linear layer. As shown in Figure \ref{different pooling}, the attentive pooling function achieves the best performance in most cases. This is likely due to the fact that the attentive pooling function is equipped with extra parameters to learn for pooling.

\begin{table}[!htb]
    \centering
    \small
    \begin{tabular}{lcccc}
        \toprule
         Model & R@1 & R@5 & R@20 \\
         \midrule
         Ours(mean pooling) & 54.72 & 78.15 & 90.21  \\
         Ours(max pooling) & 54.22 & 77.23 & 89.66 \\
         Ours(attentive pooling) & 54.79 & 78.14 & 90.25 \\
         \bottomrule
    \end{tabular}
    \caption{Experimental results of the proposed method with different pooling functions on the WikiSQL dataset.}
    \label{different pooling}
\end{table}

\subsection{Coherence Matrices of Implicit Syntactical Representations}

To investigate whether implicit syntactical representations effectively support fine-grained semantic matching akin to explicit representations, we visualized the normalized similarity matrix between implicit syntactical representations and question token embeddings (I2Q), as well as structural table representations (I2T) in Figure \ref{fig:attention_visualization}. The I2Q matrix indicates which tokens an implicit representation focuses on. Inspecting the I2Q matrix in  Figure \ref{fig:attention_visualization}(a), it is clear that $I_3$ has the closest relationship with the phrase "\textit{team}", while $I_1$ and $I_2$ both focus on the phrase "\textit{on December 5}", as there are only two phrases in the question. We observe that different implicit representations usually focus on different syntactical phrases, which is similar to explicit representations. Then, carrying different syntactical information, $I_j$ can perform neural fine-grained matching with structural header/value representations. As the I2T matrix in Figure \ref{fig:attention_visualization}(a) shows, $I_3$ matches the header "\textit{team}" with the highest score, whereas $I_1$ and $I_2$ match the header and value of "\textit{date}". Hence, the behavior of implicit representations is consistent with our design ideas. 

\end{document}